\documentclass[runningheads]{llncs}
\usepackage[T1]{fontenc}
\usepackage{graphicx,verbatim}
\usepackage{xcolor}
\usepackage{amsmath}
\usepackage{booktabs}   
\usepackage{multirow}   
\usepackage{colortbl}   
\usepackage{amssymb}
\definecolor{lightgray}{gray}{0.95}
\usepackage{hyperref}
\hypersetup{
    colorlinks=true,    
    linkcolor=blue,    
    citecolor=blue,  
    urlcolor=blue   
}
\usepackage{makecell}

\begin{document}
\title{Learning the Hierarchical Organization in Brain Network for Brain Disorder Diagnosis}\titlerunning{Learning the Hierarchical Organization in Brain Network}



%

\author{Jingfeng Tang\inst{1} \and 
Peng Cao\inst{1} \and 
Guangqi Wen\inst{2} \and 
Jinzhu Yang\inst{1} \and 
Xiaoli Liu\inst{3} \and 
Osmar R. Zaiane\inst{4}}

\authorrunning{J. Tang et al.}

\institute{Computer Science and Engineering, Northeastern University, Shenyang, China \\
\email{caopeng@cse.neu.edu.cn} \and
School of Computer Science and Artificial Intelligence, Shandong Normal University, Jinan, China \and
AiShiWeiLai AI Research, Beijing, China \and
Amii, University of Alberta, Edmonton, Alberta, Canada}
  
\maketitle              
\begin{abstract}
Brain network analysis based on functional Magnetic Resonance Imaging (fMRI) is pivotal for diagnosing brain disorders. Existing approaches typically rely on predefined functional sub-networks to construct sub-network associations. However, we identified many cross-network interaction patterns with high Pearson correlations that this strict, prior-based organization fails to capture. To overcome this limitation, we propose the Brain Hierarchical Organization Learning (BrainHO) to learn inherently hierarchical brain network dependencies based on their intrinsic features rather than predefined sub-network labels. Specifically, we design a hierarchical attention mechanism that allows the model to aggregate nodes into a hierarchical organization, effectively capturing intricate connectivity patterns at the subgraph level. 
To ensure diverse, complementary, and stable organizations, we incorporate an orthogonality constraint loss, alongside a hierarchical consistency constraint strategy, to refine node-level features using high-level graph semantics. Extensive experiments on the publicly available ABIDE and REST-meta-MDD datasets demonstrate that BrainHO not only achieves state-of-the-art classification performance but also uncovers interpretable, clinically significant biomarkers by precisely localizing disease-related sub-networks. 

\end{abstract}
\section{Introduction}

\begin{figure}
\includegraphics[width=\textwidth]{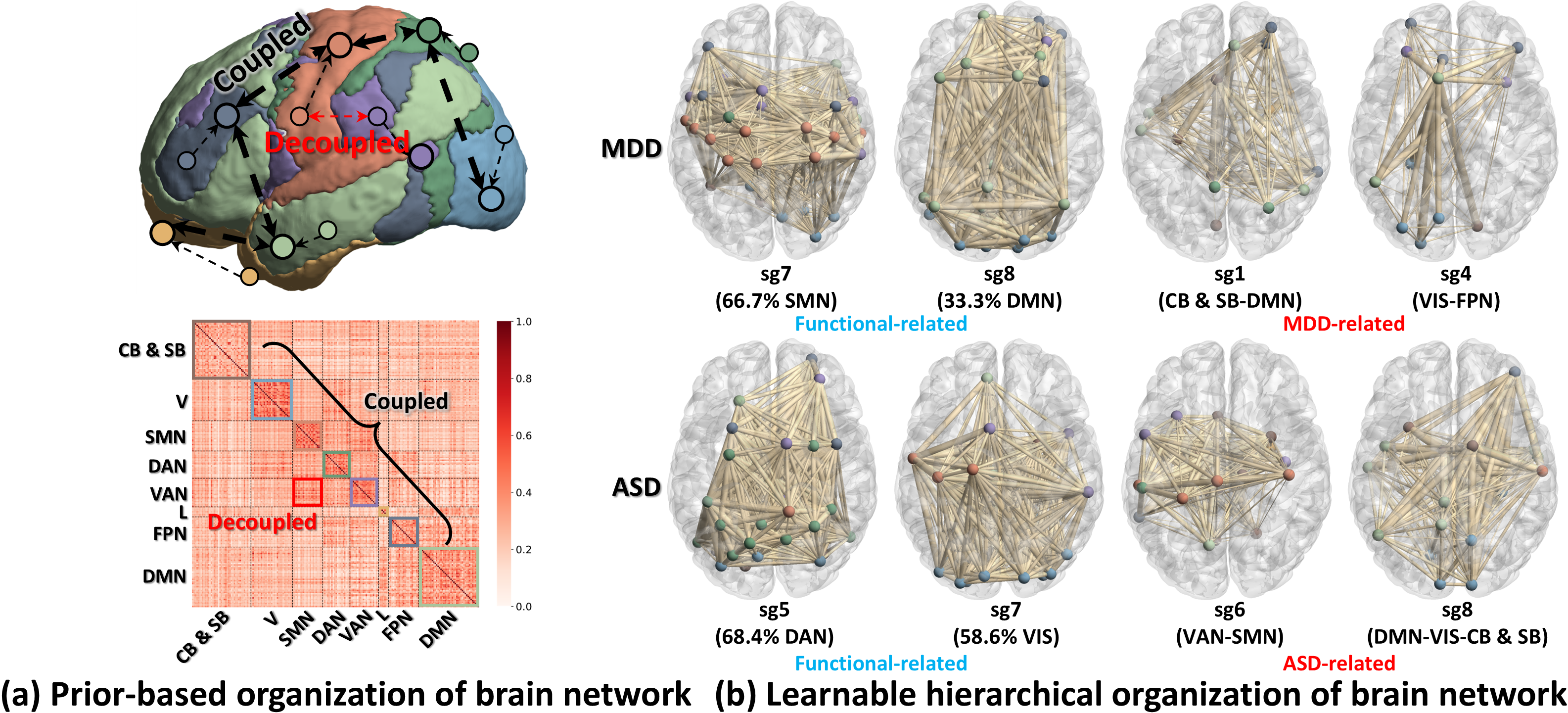}
\centering
\caption{(a) Average Pearson correlation between brain regions on the ABIDE dataset, revealing strong interactions of brain regions across predefined sub-networks. (b) Learned sub-networks on the ABIDE and REST-meta-MDD datasets. Derived from attention weights, these visualizations demonstrate that BrainHO uncovers disease-related sub-networks that span predefined sub-networks(identified via subgraph to graph attention weights) as well as functionally consistent sub-networks(determined by the proportion of functional sub-networks).} \label{fig1}
\end{figure}

Brain disorders, such as Autism Spectrum Disorder (ASD) and Major Depressive Disorder (MDD), are fundamentally underpinned by the dysfunction and atypical interactions of functional brain sub-networks\cite{padmanabhan2017default,10.1073/pnas.1900390116}.
To better analyze sub-networks, recent studies\cite{10.1007/978-3-031-43993-3_28,pei2025community}have shifted focus from node-level learning across the whole-brain network to community(subnetwork)-level learning based on existing predefined functional sub-network atlas, such as the Yeo7\cite{doi:10.1152/jn.00338.2011} and Smith10\cite{doi:10.1073/pnas.0905267106} atlas. 
Com-TF\cite{10.1007/978-3-031-43993-3_28} pioneers the use of the Yeo7\cite{doi:10.1152/jn.00338.2011} atlas to independently model intra- and inter-subnetwork interactions via local and global Transformers. Following this separate modeling paradigm, CAGT\cite{pei2025community} further integrates spatial and topological information, while DHGFormer\cite{xue2025dhgformer} leverages it to enhance GNN performance.
Existing approaches implicitly assume strict functional boundaries between predefined sub-networks. By modeling them independently, these methods fail to capture latent cross-network interactions.
However, it is worth noting that significant correlations often exist between nodes from different sub-networks, shown in Fig.\ref{fig1}(a). 
For example, the research finding \cite{10.3389/fnhum.2017.00418} reveals that individuals with ASD exhibit significantly altered functional connectivity in key regions, including the left inferior frontal gyrus (Frontal\_Inf\_Oper\_L) and the left superior temporal gyrus (Temporal\_Sup\_L). However, in fixed atlases like Yeo7, the Frontal\_Inf\_Oper\_L and Temporal\_Sup\_L are hard-coded into separate communities (e.g., the VAN and the SMN, respectively). Consequently, the vital cross-network interactions between these regions fail to be captured, which obscures critical pathological features and ultimately degrades classification accuracy.
Therefore, exploiting the hierarchical brain organization and establishing a flexible subgraph representation for distinct disorders is crucial for better understanding the underlying dysfunction in brain networks.



To address this limitation, we propose Brain Hierarchical Organization Learning (BrainHO) for modeling intrinsicly hierarchical modular organization, which circumvents the strict functional constraints, thereby emulating the human brain's underlying mechanisms\cite{10.3389/neuro.11.037.2009}.
Different from prior works\cite{10.1007/978-3-031-43993-3_28} that rely on the strict parcellation of the brain atlas, we introduce learnable subgraph tokens to dynamically aggregate information from the brain network based on node feature affinity rather than predefined labels.
To ensure that the learnable subgraphs capture diverse, complementary and stable organizations, we introduce an orthogonality constraint to encourage diverse connectivity patterns, and a hierarchical consistency strategy to refine node-level features using high-level graph semantics.

We summarize our main contributions as follows:
\begin{itemize}
     \item We propose BrainHO, a framework that breaks the limits of fixed functional atlases by modeling the hierarchical organization of brain networks in a learnable manner. This approach captures complex cross-network interactions, thereby providing interpretable insights into the underlying mechanisms of brain dysfunctions.
    
     \item We design a hierarchical attention mechanism with an orthogonality constraint to learn diverse subgraphs, coupled with a hierarchical consistency constraint mechanism to refine local node features for being highly disease-discriminative.
    
     \item Extensive experiments on the ABIDE and REST-meta-MDD datasets demonstrate that BrainHO achieves state-of-the-art performance and validates the clinical significance of the disease-related sub-networks learned by BrainHO.
\end{itemize}

\section{Method}
\subsection{overview}

\begin{figure}
\includegraphics[width=\textwidth]{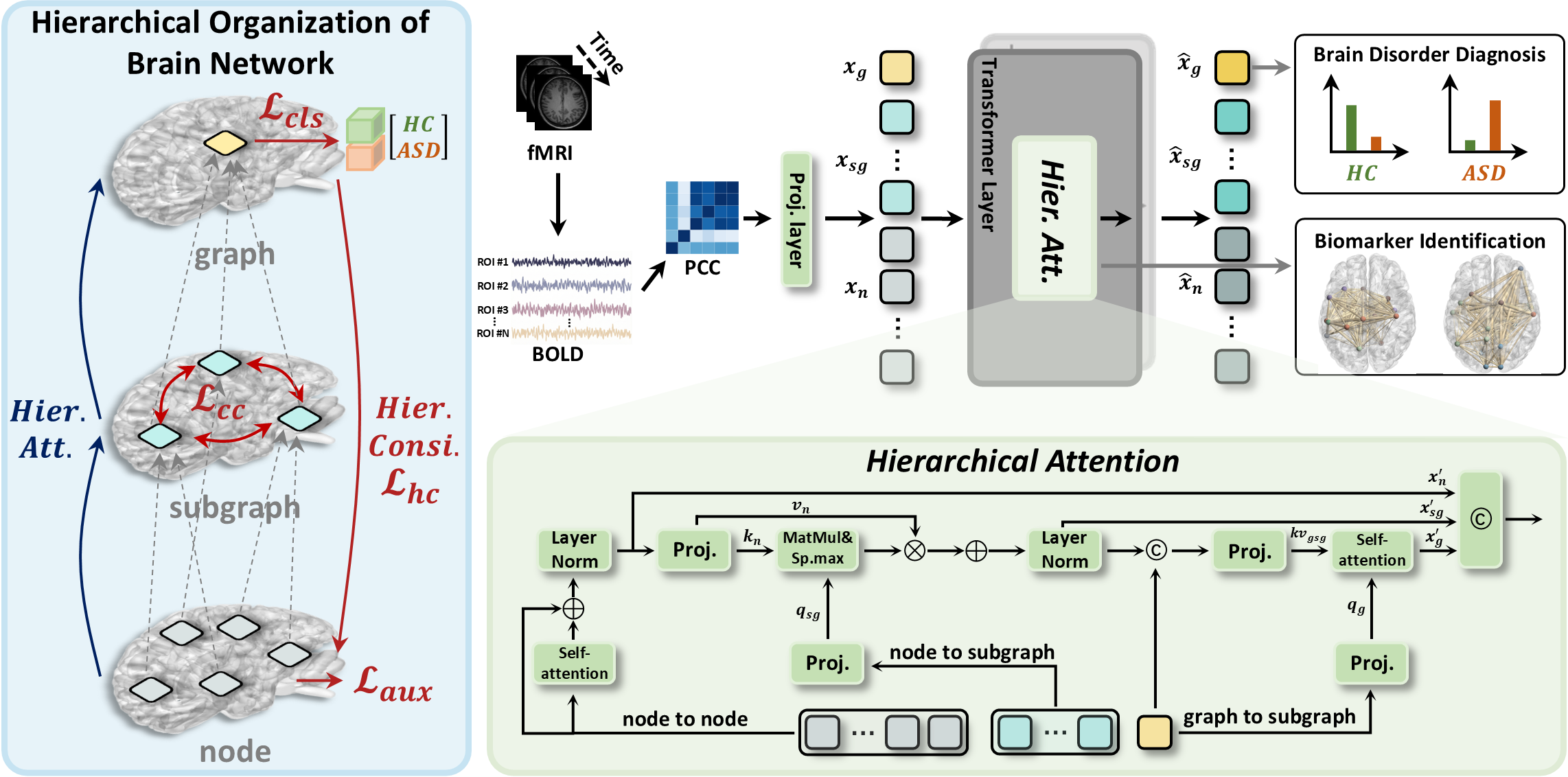}
\centering
\caption{Framework of the proposed BrainHO. The model constructs the brain network in a bottom-up attention learning manner while enforcing top-down hierarchical consistency. The Hierarchical Attention module comprises three distinct stages: node-to-node, node-to-subgraph, and subgraph-to-graph.} \label{fig2}
\end{figure}

As illustrated in Figure \ref{fig2}, BrainHO first projects the fMRI-derived Pearson Correlation Coefficient (PCC) matrix into node tokens $\mathbf{X}_n$. To capture hierarchical brain organization, a hierarchical attention mechanism dynamically aggregates these nodes along with learnable subgraph ($\mathbf{X}_{sg}$) and graph ($\mathbf{X}_g$) tokens based on node feature affinity. Concurrently, an orthogonality constraint ($\mathcal{L}_{oc}$) ensures diverse and non-redundant subgraph patterns. Finally, the final graph token $\hat{\mathbf{X}}_{g}$ drives disorder classification ($\mathcal{L}_{cls}$), while local node features are refined in parallel via a hierarchical consistency constraint ($\mathcal{L}_{hc}$) and an auxiliary classification loss ($\mathcal{L}_{aux}$).


\subsection{Hierarchical Attention Mechanism}
To effectively model the brain functional connectivity, we propose a Hierarchical Attention mechanism that organizes information at three distinct levels: node-level, subgraph-level, and graph-level. This module aggregates features in a bottom-up manner using learnable tokens.

\subsubsection{Node to Node:} The transition from the node-level to the subgraph-level involves two sequential steps: node interaction and subgraph aggregation.

First, to capture intrinsic dependencies among brain regions, we update the initial node features $\mathbf{X}_{n} \in \mathbb{R}^{N \times d}$ using a standard self-attention module, ensuring the representations incorporate global network information.

\subsubsection{Node to subgraph:} We introduce a set of learnable subgraph tokens $\mathbf{X}_{sg} \in \mathbb{R}^{K \times d}$ to represent latent sub-networks. These tokens serve as Queries to aggregate information from the updated node representations $\mathbf{X}^{\prime}_{n}$, which act as Keys and Values. To effectively filter noise and learn interpretable sub-networks, we employ the Sparsemax activation function\cite{pmlr-v48-martins16}. The updated subgraph features $\mathbf{X}^{\prime}_{sg}$ are obtained as follows:
\begin{equation}
\mathbf{A}_{(n \to sg)} = \text{Sparsemax}\left(\frac{(\mathbf{X}_{sg}\mathbf{W}_Q)(\mathbf{X}^{\prime}_{n}\mathbf{W}_K)^T}{\sqrt{d_h}}\right)
\end{equation}
\begin{equation}
\mathbf{X}^{\prime}_{sg} = \text{LN}\left(\mathbf{X}_{sg} + (\mathbf{A}_{(n \to sg)})(\mathbf{X}^{\prime}_{n}\mathbf{W}_V)\right)
\end{equation}
where $\mathbf{W}_Q, \mathbf{W}_K, \mathbf{W}_V$ are projection matrices, $\text{LN}(\cdot)$ is layer normalization, and $d_h$ is the dimension of each attention head. Unlike Softmax, Sparsemax induces a sparse probability distribution, ensuring that each subgraph token selectively attends to a relevant subset of brain nodes.

\subsubsection{Subgraph to graph:} In the final stage, we integrate the learned subgraph-level features into a holistic graph-level representation for disease diagnosis. We initialize a global graph token $\mathbf{X}_{g} \in \mathbb{R}^{1 \times d}$ to act as the Query. To allow the global token to capture both the subgraph information and its own features, the Keys and Values are projected from the concatenation of the graph token itself and the updated subgraph features $\mathbf{X}^{\prime\prime}_{sg}$.
We employ the standard self attention in this stage to densely aggregate subgraph information:
\begin{equation}
\mathbf{A}_{(sg \to g)} = \text{Softmax}\left(\frac{(\mathbf{X}_{g}\mathbf{W}'_Q)(\text{Concat}(\mathbf{X}_{g}, \mathbf{X}^{\prime\prime}_{sg})\mathbf{W}'_K)^T}{\sqrt{d_h}}\right)
\end{equation}
This process generates a comprehensive graph-level embedding $\mathbf{X}^{\prime}_{g}$ that encapsulates the interactions among different functional sub-networks. Subsequently, $\mathbf{X}^{\prime}_{g}$ is concatenated with the updated subgraph tokens $\mathbf{X}^{\prime}_{sg}$ and node tokens $\mathbf{X}^{\prime}_{n}$, and fed into the Feed-Forward Network for classification.

\subsection{Subgraph Orthogonality} 
To encourage diversity and avoid redundancy among the learnable subgraph tokens, we impose a subgraph orthogonality constraint. Specifically, let $\hat{\mathbf{X}}_{sg} \in \mathbb{R}^{K \times d}$ denote the $L_2$-normalized subgraph tokens from the last transformer layer. We compute the pairwise similarity matrix $\mathbf{S} = \hat{\mathbf{X}}_{sg} \hat{\mathbf{X}}_{sg}^T \in \mathbb{R}^{K \times K}$, where $\mathbf{S}_{i,j}$ represents the cosine similarity between the $i$-th and $j$-th subgraph tokens. Ideally, different tokens should be orthogonal ($\mathbf{S}_{i,j} = 0$ for $i \neq j$) while maintaining self-similarity ($\mathbf{S}_{i,i} = 1$). To enforce this, we apply a cross-entropy loss over the similarity matrix, treating the diagonal indices as the ground truth:
\begin{equation}
\mathcal{L}_{\text{OC}} = - \frac{1}{K} \sum_{i=1}^{K} \log \left( \frac{\exp(\mathbf{S}_{i,i})}{\sum_{j=1}^{K} \exp(\mathbf{S}_{i,j})} \right)
\end{equation}
Minimizing $\mathcal{L}{oc}$ encourages the model to explore comprehensive and precise sub-networks connectivity patterns.

\subsection{Hierarchical Consistency}
To refine local node representations under high-level context, we introduce a hierarchical consistency strategy. We construct an auxiliary classifier that predicts disorder labels solely from node-level features. Specifically, the final graph token $\hat{\mathbf{X}}_{g}$, which aggregates information hierarchically, acts as a teacher to produce logits $z_g$. Simultaneously, the final node tokens $\hat{\mathbf{X}}_{n}$ are pooled into a representative vector to generate student logits $z_n$ via an auxiliary head. Additionally, the auxiliary head is supervised by the ground truth labels via a standard cross-entropy loss $\mathcal{L}_{aux}$. To enforce semantic consistency between the global graph context and local node representations, we minimize the Kullback-Leibler divergence:
\begin{equation}
\mathcal{L}_{hc} = \tau^2 \cdot \text{KL}\left( \text{LogSoftmax}\left(\frac{z_n}{\tau}\right) \parallel \text{Softmax}\left(\frac{z_g}{\tau}\right) \right)
\end{equation}
where $\tau$ is the temperature parameter.

The total objective function is a weighted sum of the classification loss $\mathcal{L}_{cls}$, auxiliary classification loss $\mathcal{L}_{aux}$, orthogonality loss $\mathcal{L}_{oc}$ and consistency loss $\mathcal{L}_{hc}$.

\begin{equation}
\min_{\Theta} \mathcal{L}_{total} = \mathcal{L}_{cls} + \mathcal{L}_{aux} + \alpha \mathcal{L}_{oc} + \beta \mathcal{L}_{hc}
\end{equation}
where $\Theta$ denotes the learnable parameters of the model, and $\alpha, \beta$ are the weights to balance the loss terms.

\section{Experiments}
\subsection{Dataset and experimental setup}

We utilized rs-fMRI data from ABIDE \cite{craddock2013neuro} ($N=1009$, 516 ASD/493 HCs, C-PAC preprocessed) and REST-meta-MDD \cite{10.1073/pnas.1900390116,10.1093/psyrad/kkac005} ($N=2380$, 1276 MDD/1104 HCs, DPARSF preprocessed). Both datasets were parcellated using the Craddock 200 atlas \cite{craddock2012whole} to extract node features.

We employed 5-fold stratified cross-validation, reserving 25\% of training samples as a validation set for early stopping. We report the mean and standard deviation of Accuracy (ACC), AUC, Sensitivity (SEN), and Specificity (SPE) on the test set.


Implemented in PyTorch (A100 GPU), our framework utilized two 8-head transformer layers trained for 200 epochs (batch size 64) using AdamW (LR=$1\text{e-}4$, weight decay=$1\text{e-}4$) and CosineAnnealingLR (min LR=$1\text{e-}5$). We set $\alpha=1.3$, $K=8$, $d=384$, $d_h=d/8$, $\tau=2.0$, and applied a sigmoid-scheduled $\beta$ (weight 0.2, center $1/4$ steps, slope 0.001). Following prior works\cite{NEURIPS2022_a408234a,xue2025dhgformer}, Mixup\cite{zhang2018mixup} augmentation was adopted.

\subsection{Results and Discussion}
\subsubsection{Comparison with state-of-the-art Methods:}

As shown in Table \ref{tab1}, BrainHO outperformed state-of-the-art baselines on the ABIDE dataset, achieving the highest accuracy ($69.68\%$) and AUC ($73.80\%$). 
It is noteworthy that while advanced methods such as ALTER\cite{NEURIPS2024_2bd3ffba}, DHGFormer\cite{xue2025dhgformer}, and LHDFormer\cite{XueRun_Adaptive_MICCAI2025} incorporate raw BOLD signals ($\mathcal{B}$) as additional temporal input, BrainHO achieves superior performance relying solely on the static PCC connectivity matrix ($\mathcal{P}$).
This result highlights the capability of our framework to extract diease features without complex temporal inputs, validating the superiority of hierarchical subgraph learning over flat node-level modeling approaches(e.g., BNT\cite{NEURIPS2022_a408234a}, ALTER\cite{NEURIPS2024_2bd3ffba}).
Furthermore, BrainHO attained the highest sensitivity ($73.11\%$), significantly exceeding the second-best Com-BrainTF\cite{10.1007/978-3-031-43993-3_28}. Unlike LHDFormer\cite{XueRun_Adaptive_MICCAI2025}, which sacrifices sensitivity for specificity, BrainHO maintained a balanced performance, effectively capturing latent cross-subnetwork interactions. Regarding the larger-scale REST-meta-MDD dataset, BrainHO demonstrated robust generalization, securing the highest Accuracy ($64.71\%$) and Sensitivity ($67.43\%$). Although LHDFormer exhibited a marginally higher AUC, BrainHO prioritized the identification of positive cases without compromising overall accuracy. This balance makes BrainHO more clinically suitable for effective disease screening compared to baselines that exhibit trade-offs between accuracy and sensitivity.

\begin{table}[htbp]
    \centering
    \caption{Performance comparison on ABIDE and REST-meta-MDD datasets. The Input column specifies the data type used by each method, where $\mathcal{B}$ represents raw BOLD signals and $\mathcal{P}$ denotes the static PCC connectivity matrix.}
    \label{tab1}
    \begin{tabular}{c|lccccc}
        \toprule
        Dataset & Methods & Input & ACC(\%) & AUC(\%) & SEN(\%) & SPE(\%) \\
        \midrule
        \multirow{6}{*}{\makecell{REST-\\meta-MDD}} 
        & BNT & $\mathcal{P}$ & 63.28$\pm$1.48 & 69.08$\pm$1.49 & 65.03$\pm$4.83 & 61.41$\pm$6.48 \\
        & Com-BrainTF & $\mathcal{P}$ &  63.24$\pm$2.34 & 68.73$\pm$1.76 & 60.93$\pm$7.67 & 66.21$\pm$6.90 \\
        & ALTER & $\mathcal{B},\mathcal{P}$ &  63.95$\pm$0.72 & 68.91$\pm$0.88 & 64.94$\pm$6.28 & 62.75$\pm$7.75 \\
        & DHGFormer & $\mathcal{B},\mathcal{P}$ &  61.60$\pm$1.36 & 65.18$\pm$1.79 & 64.73$\pm$3.55 & 58.01$\pm$3.84 \\
        & LHDFormer & $\mathcal{B},\mathcal{P}$ &  63.95$\pm$1.88 & \textbf{69.96$\pm$1.80} & 65.08$\pm$5.25 & \textbf{62.79$\pm$8.36} \\
        \rowcolor{gray!15} \cellcolor{white} 
        & BrainHO(Ours) & $\mathcal{P}$ & \textbf{64.71$\pm$2.01} & 68.83$\pm$1.34 & \textbf{67.43$\pm$4.19} & 61.51$\pm$3.65 \\
        \midrule
 \multirow{6}{*}{ABIDE} 
        & BNT & $\mathcal{P}$ & 66.61$\pm$2.47 & 73.32$\pm$2.48 & 67.18$\pm$6.34 & 66.08$\pm$7.61 \\
        & Com-BrainTF & $\mathcal{P}$ & 67.69$\pm$3.09 & 73.20$\pm$3.40 & 69.28$\pm$10.22 & 65.78$\pm$5.56 \\
        & ALTER & $\mathcal{B},\mathcal{P}$ & 66.40$\pm$1.73 & 73.46$\pm$2.22 & 66.17$\pm$6.44 & 67.33$\pm$5.46 \\
        & DHGFormer & $\mathcal{B},\mathcal{P}$ & 65.22$\pm$3.50 & 69.15$\pm$3.51 & 67.62$\pm$7.03 & 63.15$\pm$7.85 \\
        & LHDFormer & $\mathcal{B},\mathcal{P}$ & 67.50$\pm$1.83 & 73.11$\pm$2.10 & 63.82$\pm$6.10 & \textbf{71.87$\pm$7.47} \\
        \rowcolor{gray!15} \cellcolor{white} 
        & BrainHO(Ours) & $\mathcal{P}$ & \textbf{69.68$\pm$2.11} & \textbf{73.80$\pm$2.52} & \textbf{73.11$\pm$6.73} & 67.08$\pm$4.17 \\
 \midrule
 \multirow{6}{*}{ABIDE} 
&Baseline & $\mathcal{P}$ & 61.55$\pm$2.95 & 68.85$\pm$2.91 & 62.11$\pm$13.20 & 62.05$\pm$12.69 \\
&w/o HA & $\mathcal{P}$           & 61.74$\pm$1.78 & 69.09$\pm$1.26 & 59.57$\pm$9.02 & 64.07$\pm$6.99 \\
&w/o $\mathcal{L}_{oc}$ & $\mathcal{P}$    & 65.61$\pm$1.46 & 71.22$\pm$1.49 & 73.10$\pm$11.28 & 59.19$\pm$10.74 \\
&w/o $\mathcal{L}_{aux}$  & $\mathcal{P}$    & 67.20$\pm$2.13 & 72.24$\pm$3.56 & 69.32$\pm$5.17 & 65.39$\pm$4.98 \\
&w/o $\mathcal{L}_{hc}$  & $\mathcal{P}$    & 68.89$\pm$3.05 & 73.67$\pm$2.60 & 67.69$\pm$5.12 & 70.58$\pm$2.71 \\
        \bottomrule
    \end{tabular}
\end{table}

\subsubsection{Ablation study:} 
We validated BrainHO via an ablation study (Table \ref{tab1}).
The Baseline (standard Transformer) and w/o HA variant (where HA is replaced by an attention module) yielded only $61.55\%$ and $61.74\%$ accuracy, significantly underperforming the full model and indicating standard attention fails to capture brain hierarchies.
Furthermore, removing $\mathcal{L}_{oc}$ led to a sharp accuracy drop to $65.61\%$, confirming that regularizing subgraph orthogonality is critical for preventing feature collapse and ensuring diverse connectivity patterns. The removal of the auxiliary supervision ($\mathcal{L}_{aux}$) resulted in a performance decline to 67.20\%. This suggests that explicit supervision on local representations is indispensable for ensuring the distinctiveness of node-level features. Similarly, excluding $\mathcal{L}_{hc}$ reduced accuracy to $68.89\%$, highlighting the contribution of the global-to-local (from top to bottom) in refining node features.

\subsubsection{Interpretability of the learned sub-networks:}

\begin{figure}
\includegraphics[width=\textwidth]{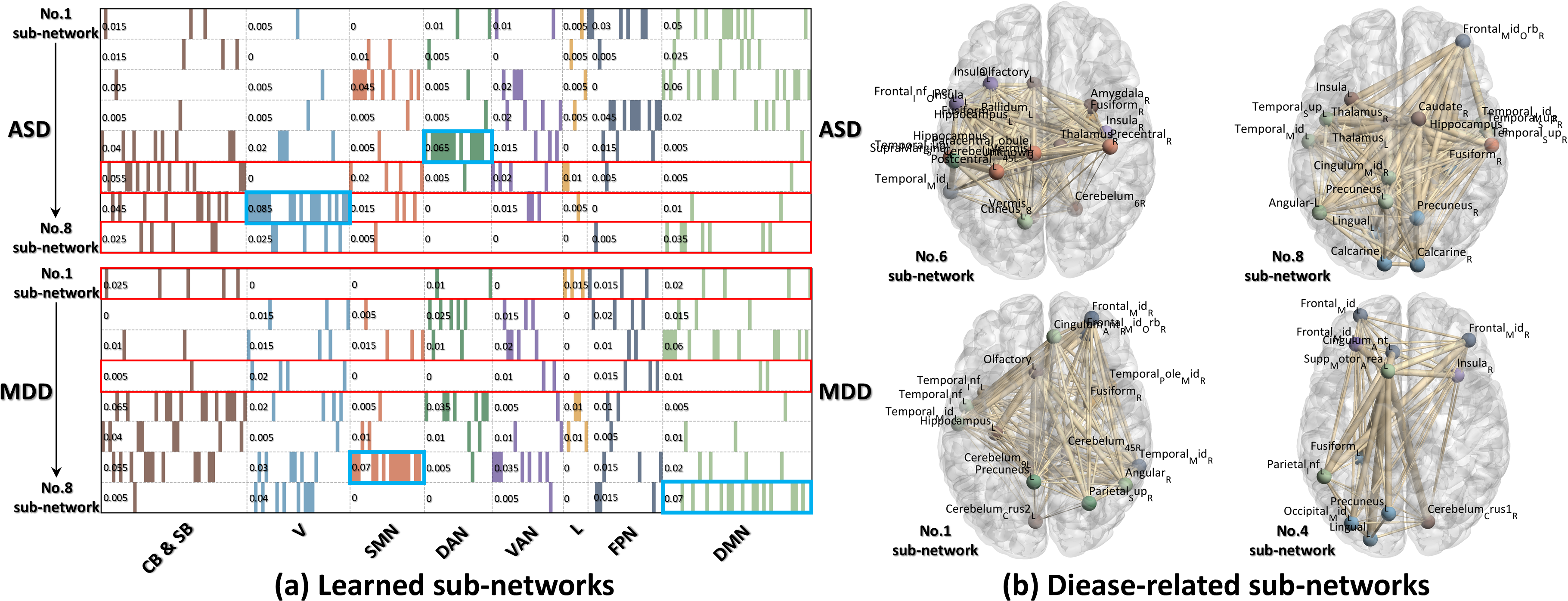}
\centering
\caption{The interpretibility of the  sub-networks identified by our model.}\label{fig3}
\end{figure}

Figure \ref{fig3}(a) shows the mapping from the learned sub-networks (rows) to predefined functional networks (columns). Notably, certain learned sub-networks predominantly align with the defined sub-networks. For example, the No. 7 sub-network for MDD and the No. 5 sub-network for ASD substantially overlap with the SMN and DAN, covering 66.7\% and 68.4\% of their nodes, respectively. This proves that BrainHO can uncover functionally consistent sub-networks (Shown in Figure \ref{fig1}(b)) consistent with the prior atlas. 
Moreover, other identified sub-networks flexibly encompass brain regions from multiple predefined functional networks, effectively capture the latent cross-network interactions. This distinct and complementary distribution underscores the orthogonality of the learned sub-networks, precisely enabling precise localization of disease-related cross-network anomalies critical for diagnosis.

Since the CC200 atlas lacks explicit labels, Figure \ref{fig3}(b) visualizes the disease-related sub-networks using the mapped AAL116 names for better interpretation. For ASD, the No. 6 sub-network identifies key nodes like Fusiform\_L, Amygdala\_R\cite{10.1002/hbm.26141}, alongside Thalamus\_R, Fusiform\_R, Vermis, Cerebelum\_4-\_5\_L\cite{10.1093/cercor/bhad451}. Additionally, the No. 8 sub-network highlights Temporal\_Sup and Temporal\_Mid, which are crucial for ASD-related social perception \cite{10.1093/brain/awh404,JOU2010205}. For MDD, the No. 1 sub-network effectively captures the Hippocampus and Cingulum\_Ant, both of which are fundamental to emotion and executive function regulation \cite{schmaal2016subcortical,pizzagalli2011frontocingulate}. 
NO. 4 sub-network includes these nodes (Frontal\_Mid\_R, Cingulum\_Ant\_L, Parietal\_Inf\_L, Insula\_R, Frontal\_Mid\_L, and Precuneus\_L), and their functional abnormalities are closely associated with MDD\cite{10.3389/fnhum.2014.00692}.


\section{Conclusion}
In this paper, we proposed the Brain Hierarchical Organization Learning framework (BrainHO) to address the limitations of strict, predefined functional atlases in brain network analysis. BrainHO dynamically aggregates brain regions based on intrinsic functional affinity rather than predefined sub-network labels. We introduced a hierarchical attention mechanism coupled with an orthogonality constraint to ensure diverse sub-network representations, alongside a hierarchical consistency constraint strategy to refine local features using global semantics. Extensive experiments on the ABIDE and REST-meta-MDD datasets demonstrated that BrainHO achieved state-of-the-art classification performance. Furthermore, the visualization of learned organization confirms the model's capacity to uncover interpretable, clinically significant biomarkers, providing new insights into the hierarchical neuropathology of brain disorders.

\bibliographystyle{splncs04}
\bibliography{ref}
%




\end{document}